%% file: main.tex
\pdfoutput=1

\documentclass[11pt]{article}

\usepackage[final]{acl}

\usepackage{times}
\usepackage{latexsym}

\usepackage[T1]{fontenc}

\usepackage[utf8]{inputenc}

\usepackage{microtype}

\usepackage{inconsolata}

\usepackage{graphicx}

\usepackage{forest} 
\usetikzlibrary{arrows.meta, shapes, positioning, shadows, trees}
\usepackage{pifont} 
\usepackage{booktabs} 
\usepackage{enumitem} 
\usepackage{tablefootnote} 
\usepackage{url} 
\usepackage{multirow}   
\usepackage{makecell}   
\usepackage[table]{xcolor} 

\newcommand{\clickablecheckmark}[1]{\mbox{\href{#1}{\ding{51}}}}

\title{A Survey of Deep Learning for Geometry Problem Solving}

\author{Jianzhe Ma$^1$
~~~Wenxuan Wang$^{1*}$
~~~Qin Jin$^1$\thanks{Qin Jin and Wenxuan Wang are corresponding authors.} \\
$^1$Renmin University of China\\
\texttt{\{majianzhe, wangwenxuan, qjin\}@ruc.edu.cn}
}

\begin{document}
\maketitle
\begin{abstract}
Geometry problem solving, a crucial aspect of mathematical reasoning, is vital across various domains, including education, the assessment of AI's mathematical abilities, and multimodal capability evaluation. The recent surge in deep learning technologies, particularly the emergence of multimodal large language models, has significantly accelerated research in this area. This paper presents a survey of the applications of deep learning in geometry problem solving, including (i) a comprehensive summary of the relevant tasks in geometry problem solving; (ii) a thorough review of related deep learning methods; (iii) a detailed analysis of evaluation metrics and methods; and (iv) a critical discussion of state-of-the-art performance, existing challenges, and promising future directions. Our objective is to offer a comprehensive and practical reference of deep learning for geometry problem solving, thereby fostering further advancements in this field. We maintain a list of relevant papers: \textit{\url{https://github.com/majianz/dl4gps}}.

\end{abstract}

\input{section/intro}
\input{section/task}
\input{section/method}

\input{section/eval}

\input{section/discuss}
\input{section/conclu}

\section*{Limitations}
Our survey focuses on the intersection of deep learning and GPS tasks in the past decade, and may not fully represent the development process of the entire field. In addition, given the rapid development of this field, our survey may not timely reflect the latest developments and progress before and after the survey. Furthermore, our survey is mainly dedicated to summarizing existing research work, and there are limitations in experimental analysis. Despite these limitations, this survey still provides a valuable overview of the current status and main trends in the field of deep learning for GPS, which is expected to provide a useful reference for researchers and practitioners in this field.

\section*{Acknowledgments}

We thank all reviewers for their insightful comments and suggestions. This work was partially supported by the Beijing Natural Science Foundation (No. L233008).

\bibliography{custom}


\clearpage
\twocolumn[\vspace*{2em}]

\appendix

\begin{figure}[t]
    \centering
  \includegraphics[width=0.95\columnwidth]{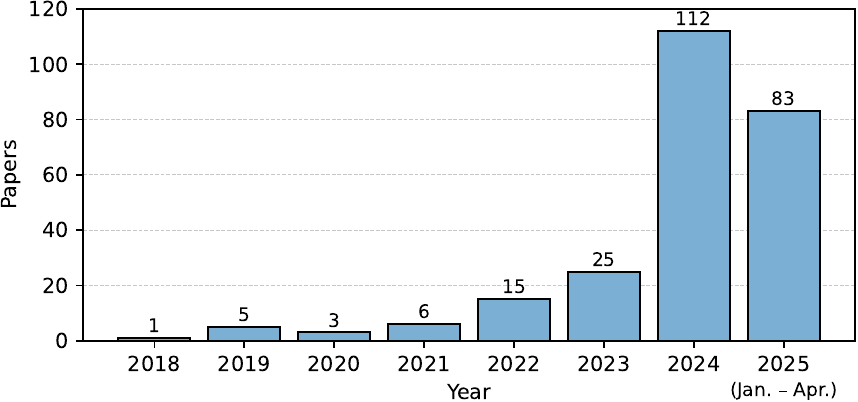}
  \caption{Papers on deep learning for geometry problem solving over the years (data for 2025 is up to April).}
  \label{fig:year}
  \vspace{-0.5cm}
\end{figure}

\input{section/data}
\input{section/other}
\input{section/encoder-decoder}


\begin{table*}[]
\centering
\fontsize{6.7pt}{8.8pt}\selectfont
\begin{tabular}{lccccccc}
\toprule
\textbf{Paper}          & \textbf{Task} & \textbf{Network}           & \textbf{Text Encoder}                                        & \textbf{Diagram Encoder}                                          & \textbf{Fusion Module}                & \textbf{Decoder}         & \textbf{Knowledge Module} \\
\midrule
\multicolumn{8}{c}{\cellcolor{gray!10}\textit{\textbf{Fundamental Tasks}}} \\
\midrule
\textbf{RSP}~\citeyearpar{RSP123}            & SP            & BiLSTM                     & BiLSTM                                                       & -                                                                 & -                                        & -                        & -                         \\
\textbf{2StepMemory}~\citeyearpar{two-step}    & SP            & attention                  & attention                                                    & -                                                                 & -                                        & -                        & -                         \\
\textbf{GIRTOOLS}~\citeyearpar{girtools}       & ER            & VGG16                      & -                                                            & -                                                                 & -                                        & -                        & -                         \\
\textbf{Arsenal}~\citeyearpar{Arsenal}        & SP            & Seq2Seq                    & RNN                                                          & -                                                                 & -                                        & -$^{\dag}$                       & -                         \\
\bottomrule
\end{tabular}
\caption{A summarization of deep learning architectures for geometry problem solving system. Task: DP: geometric diagram parsing, SP: semantic parsing for geometry problem texts, ER: geometric element recognition, DP: geometric diagram parsing, SR: geometric structure recognition, RE: geometric relation extraction, TP: geometry theorem proving, NC: geometric numerical calculation. $^{\dag}$ indicates the presence of the attention mechanism.}
\label{network_table}
\end{table*}

\end{document}

%% file: section/intro.tex
\section{Introduction}
 
As a core aspect of mathematical reasoning, \textbf{Geometry Problem Solving} (\textbf{GPS}) has long been closely tied to education and the assessment of mathematical proficiency in Artificial Intelligence (AI) systems~\citep{narboux2018computer}. 
Dating back to the 1960s, GPS is one of the earliest research areas in automated mathematical reasoning~\citep{gelernter1960empirical}.
Given the inherent connection between geometry problems and diagrams, GPS has naturally emerged as a representative multimodal mathematical task.
Solving geometry problems in the format of educational exams requires AI systems not only to interpret geometric diagrams but also to perform robust logical reasoning and numerical computation, making it an ideal benchmark for assessing perception and reasoning in deep learning models. 

Early deep learning applications primarily focused on combining deep learning modules with symbolic systems to solve geometry problems using formal languages, with the deep learning module acting as part of the overall system~\citep{aifu, inter-gps}. Neural-symbolic methods have gradually become the mainstream approach in deep learning for GPS. In recent years, work solely employing deep learning methods has also increased~\citep{eagle, gllava}. 
The rise of Multimodal Large Language Models (MLLMs), in particular, has further advanced this field, showcasing the great potential of deep learning in complex visual understanding and reasoning tasks. The number of papers on deep learning for GPS has grown rapidly, from just 1 in 2018 to 112 in 2024, and continues to increase in 2025 (see Figure~\ref{fig:year} in the Appendix).

\begin{figure}[t]
    \centering
  \includegraphics[width=\columnwidth]{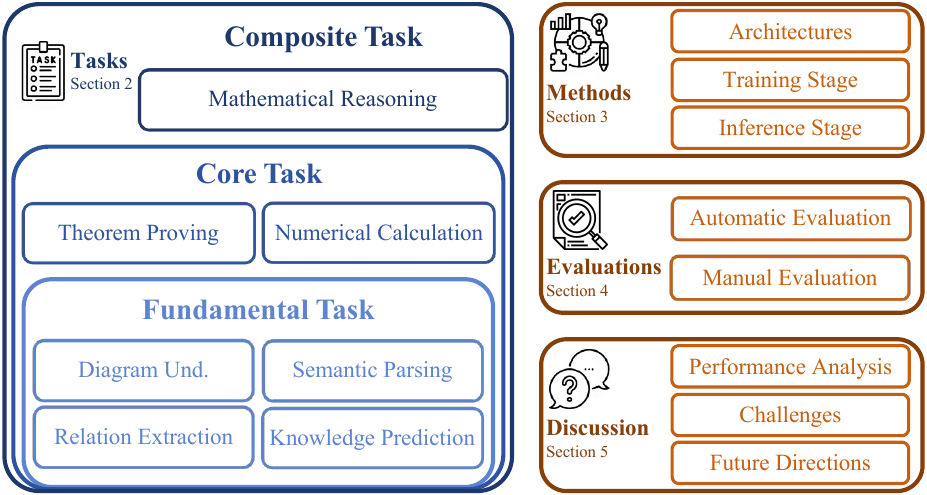}
  \caption{Overview of the survey's structure.}
    \vspace{-0.5cm}
  \label{fig:framework}
\end{figure}

Although many surveys have reviewed deep learning methods and Large Language Models (LLMs) in the broader field of mathematical reasoning~\citep{dl4math_survey, llm_for_math_survey, auto_mps_survey, mllm4math_survey}, the subfield of GPS remains underexplored compared to other mathematical areas~\citep{mwps_survey2, dl4tp}.
Recent surveys on GPS are relatively limited in scope---either concentrating solely on multimodal plane geometry problems~\citep{pgps_survey}, or lacking a comprehensive summary of relevant datasets and deep learning methods~\citep{GPS_LM_Survey}.

\tikzset{
    basic/.style  = {draw,rectangle, font=\scriptsize},
    root/.style   = {basic, rounded corners=3pt, thin, align=center, fill=white, text width=1.2cm,font=\scriptsize\bfseries},
    onode/.style = {basic, thin, rounded corners=3pt, align=left, fill=white, text width=1.2cm},
    tnode/.style = {basic, thin, rounded corners=3pt, align=left, fill=white, text width=2.5cm},
    x1node/.style = {basic, thin, rounded corners=3pt, align=left, fill=blue!5, text width=8.6cm},
    x2node/.style = {basic, thin, rounded corners=3pt, align=left, fill=blue!5, text width=5.5cm}
}
\begin{figure*}[t] 
    \centering 
    \begin{forest} for tree={
      grow=east,
      anchor=west,
      s sep+=-5pt,
      edge={-},
      edge path={
        \noexpand\path [draw, ->, \forestoption{edge}] (!u.parent anchor) -- +(5pt,0) |- (.child anchor)\forestoption{edge label};
      },
      parent anchor=east,
      child anchor=west,
    }
    [Tasks and Datasets(§\ref{task}), root
        [Composite Tasks(§\ref{adv}), onode
            [Mathematical Reasoning, tnode
                [{E.g., MATH~\citeyearpar{MATH_AMPS}, MathVista~\citeyearpar{mathvista}, Math-Vision~\citeyearpar{math-vision}, MathVerse~\citeyearpar{mathverse}}, x1node]
            ] 
        ]
        [Core Tasks(§\ref{core}), onode
            [Numerical Calculation, tnode
                [{E.g., GEOS~\citeyearpar{geos}, GeoQA~\citeyearpar{geoqa_ngs}, Geometry3K~\citeyearpar{inter-gps}, PGPS9K~\citeyearpar{PGPSNet123}}, x1node
                ]
            ] 
            [Theorem Proving, tnode
                [{E.g., Proving2H~\citeyearpar{diagramet_paradigm}, UniGeo~\citeyearpar{unigeo}, IMO-AG-30~\citeyearpar{alphageometry}, MO-TG-225~\citeyearpar{tonggeometry}}, x1node]
            ]
        ]
        [Fundamental Tasks(§\ref{fund}), onode
            [Knowledge Prediction, tnode
                [{E.g., GeoSense~\citep{geosense}, GNS-260K~\citep{gns}}, x1node]
            ]
            [Relation Extraction, tnode
                [{E.g., GeoC50~\citep{GeoC50}, GeoRE~\citep{geore}}, x1node]
            ]
            [Semantic Parsing, tnode
                [{E.g., RSP~\citep{RSP123}, Arsenal~\citep{Arsenal}, 2StepMemory~\citeyearpar{two-step}}, x1node]
            ]
            [Diagram Understanding, tnode
                [{E.g., Tangram~\citeyearpar{tangram}, Geoclidean~\citeyearpar{geoclidean}, PGDP5K~\citeyearpar{pgdp5k}, AutoGeo~\citeyearpar{autogeo}}, x1node]
                ]
            ]
        ]
    ]
    \end{forest}
    \caption{Taxonomy of tasks and datasets for geometry problem solving.} 
    \label{task_and_data_figure}
\end{figure*}

In this study, we began with several classic papers in this field, conducted a single round of forward and backward snowballing, searched Google Scholar with the keyword ``geometry'', and manually screened to ensure the relevance of the papers. As a result, we collected more than \textbf{310} academic papers that involved deep learning for GPS, and conducted a comprehensive and in-depth survey.

In the following sections, we will first summarize the tasks related to GPS in depth (§\ref{task}). Then, we comprehensively review the various methods used in the field of GPS (§\ref{method}). 
After that, we perform a systematic analysis of the evaluation metrics and methods (§\ref{eval}).
Finally, we analyze the performance of representative models, discuss the current challenges facing this field, and look forward to future development directions (§\ref{discuss}).
The survey's organizational structure is illustrated in Figure~\ref{fig:framework}.

%% file: section/task.tex
\section{Geometry Problem Solving Tasks}
\label{task}

In this section, we outline the tasks related to GPS, which are categorized into fundamental, core, and composite tasks. Fundamental tasks cover the basic abilities required for solving geometry problems, core tasks are directly tied to GPS, and composite tasks treat GPS as part of broader complex tasks. The taxonomy of tasks and datasets is shown in Figure~\ref{task_and_data_figure}, and a detailed summary of the datasets can be found in Table~\ref{basic_core_data_table} and Table~\ref{Composite_table}.

\subsection{Fundamental Tasks}
\label{fund}
In order to solve geometry problems, a deep learning system must first have a variety of fundamental capabilities, including understanding geometric diagrams, semantic parsing of geometry problem texts, extraction of geometric relationships, and prediction of geometric knowledge.

\noindent\textbf{Geometric Diagram Understanding.} 
Geometric diagram understanding aims to fully understand the information in geometric diagrams. 
It consists of multiple subtasks at different levels.
First, detect and identify basic geometric elements (such as points, lines, angles, and polygons) and their attributes (such as quantity and size)~\citep{overlapped, Retrieving_from_hand-drawn_diagrams, girtools}. This task is called \textit{Geometric Element Recognition}.
Second, based on the recognition of geometric elements, further identify and construct the structure and spatial relationship between elements~\citep{diagramet_paradigm, Algebraic_Problems_with_Geometry_Diagrams}, namely \textit{Geometric Structure Recognition}. 
These two tasks are often jointly considered as \textit{Geometric Perception} tasks~\citep{visonlyqa, gepbench}.
Third, based on geometric perception capabilities, generate formal language for geometric diagrams~\citep{pgdp5k, sp-1}. This task is also known as \textit{Geometric Diagram Parsing}. Finally, some studies use natural language to provide an accurate description of geometric diagrams. These descriptions are either generated based on diagram parsing or directly generated from geometric diagrams~\citep{gold, autogeo}, which is referred to as \textit{Geometric Diagram Captioning}.

\noindent\textbf{Semantic Parsing} for geometry problem texts. Semantic parsing is essential for converting problem text into machine-readable formal statements~\citep{End-to-end_Math_Problem_Solving}, and was a core component of early deep learning frameworks for GPS~\citep{RSP123, Neural_Semantic_Parser}. Geometry problem texts often contain multiple sentences and complex geometric information, making cross-sentence references and domain-specific content challenging~\citep{EUCLID_Tree_transducers}. Some studies also integrate diagram parsing with semantic parsing, aiming to achieve the joint parsing of text and diagrams~\citep{METEOR, Formal_Language_Generation}.

\noindent\textbf{Geometric Relation Extraction.} 
Geometric relation extraction is a well-defined task that involves extracting geometric relationships either from the question text~\citep{Uniform_Vectorized_Syntax-Semantics}, or jointly from both text and diagrams~\citep{GeoC50}, and representing them in structured formats such as triples~\citep{BiLSTM-CRF} or knowledge graphs~\citep{sgko}.
The model achieves a deep understanding of the problem by extracting geometric relationships in geometry problems rather than using natural language~\citep{Syntax-Semantics, Supervised_Approach}.

\noindent\textbf{Geometric Knowledge Prediction.}
Geometric knowledge prediction aims to evaluate the model's understanding of geometry by predicting the geometric principles~\citep{geosense} and theorems~\citep{inter-gps} (i.e., geometric knowledge) required to solve geometry problems~\citep{gns}. The model needs to predict the relevant geometric knowledge required to solve the problem based on the input question and apply it in the reasoning process~\citep{e-gps}.

\subsection{Core Tasks}
\label{core}

\begin{figure}[t]
    \centering
  \includegraphics[width=\columnwidth]{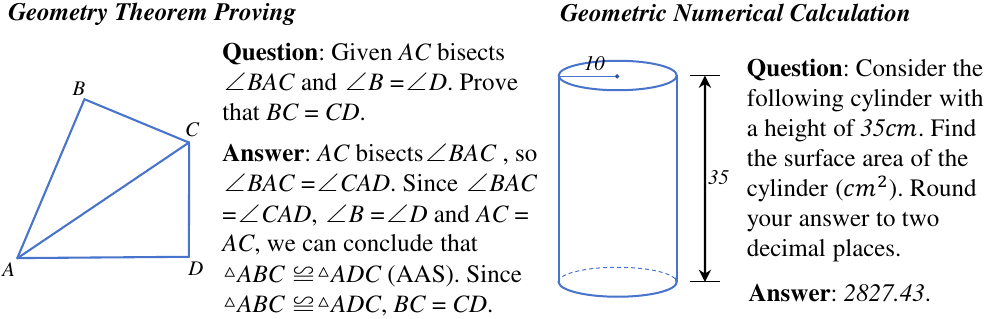}
  \caption{An example of geometry theorem proving and geometric numerical calculation problem.}
    \vspace{-0.5cm}
  \label{fig:gps_example}
\end{figure}

GPS can be categorized into geometry theorem proving and geometric numerical calculation~\citep{unigeo}. 
Building upon the capabilities established in the fundamental tasks, the model needs to solve geometry problems in the format of educational exams.
See Figure~\ref{fig:gps_example} for an example.

\noindent\textbf{Geometry Theorem Proving.}
Geometry theorem proving is a long-standing task in the field of AI~\citep{gelernter1960empirical, kapur1986using}. The input is a geometry theorem that requires proof, and the goal is to output a detailed derivation process of the proof, usually focusing on plane geometry.

\noindent\textbf{Geometric Numerical Calculation.}
Geometric numerical calculation has gradually emerged with the introduction of new datasets in recent years~\citep{geos, geos++}. 
The input is a geometry problem involving the calculation of a certain geometric value (such as length or angle), and the desired output is a concise answer to the problem, without necessarily providing a complete reasoning process.
Its question types can usually be divided into several categories, including plane geometry, solid geometry, and analytic geometry.

\subsection{Composite Tasks}
\label{adv}
Recently, GPS has also often appeared as a subtask of composite tasks, mainly used to explore the model's ability in mathematical reasoning.

\noindent\textbf{Mathematical Reasoning.}
Geometry is an important part of mathematics, and geometric diagrams are also a typical type of mathematical synthetic images. Therefore, geometry problems are often included in single-modal or multi-modal mathematical benchmarks~\citep{MATH_AMPS, mathvista} to evaluate the performance of models in mathematical reasoning tasks. 

Besides the GPS tasks above, we also summarize other geometry-related tasks sharing similar fundamental tasks in Appendix~\ref{other}, along with a detailed summary of their datasets in Table~\ref{other_task_table}.

%% file: section/method.tex
\section{Methods for Geometry Problem Solving}
\label{method}
In this section, we comprehensively review deep learning methods for GPS. We first introduce the relevant architectures, then classify and summarize other methods according to the training and inference stages. The taxonomy of these methods is shown in Figure~\ref{method_figure}.

\tikzset{
    basic/.style  = {draw,rectangle, font=\scriptsize},
    root/.style   = {basic, rounded corners=3pt, thin, align=center, fill=white,text width=1.2cm,font=\scriptsize\bfseries},
    onode/.style = {basic, thin, rounded corners=3pt, align=left, fill=white, text width=1.5cm},
    tnode/.style = {basic, thin, rounded corners=3pt, align=left, fill=white, text width=2.5cm},
    t1node/.style = {basic, thin, rounded corners=3pt, align=left, fill=white, text width=2cm},
    x1node/.style = {basic, thin, rounded corners=3pt, align=left, fill=blue!5, text width=8.4cm},
    x2node/.style = {basic, thin, rounded corners=3pt, align=left, fill=blue!5, text width=5.65cm}
}

\begin{figure*}[h!] 
    \centering 
    \begin{forest} for tree={
      grow=east,
      anchor=west,
      s sep+=-5pt,
      edge={-},
      edge path={
        \noexpand\path [draw, ->, \forestoption{edge}] (!u.parent anchor) -- +(5pt,0) |- (.child anchor)\forestoption{edge label};
      },
      parent anchor=east,
      child anchor=west,
    }
    [Methods(§\ref{method}), root
        [Inference(§\ref{infer}), onode
            [Knowledge-Augmented, tnode
                [Others, t1node
                    [{E.g., Learning to Plan~\citep{learning2plan}, EM$^2$~\citeyearpar{em2}}, x2node,
                    ]
                ]
                [Visual Aids, t1node
                    [{E.g., VAP~\citep{Vision-Augmented_Prompting}, VisuoThink~\citeyearpar{visuothink}}, x2node
                    ]
                ]
                [Few-shot Learning, t1node
                    [{E.g., ICL~\citep{cds}, RAG~\citep{geocoder}}, x2node
                    ]
                ]
            ]
            [Test-Time Scaling, tnode
                [Others, t1node
                    [{E.g., $R^3V$~\citep{r3v}, CEO~\citep{jin2025TTSforMAS}}, x2node,
                    ]
                ]
                [Verification, t1node
                    [{E.g., PRM~\citeyearpar{atomthink}, LECO~\citeyearpar{leco}, GenRM~\citeyearpar{genrm}}, x2node,
                    edge path={
                        \noexpand\path[\forestoption{edge},-, >={latex}] 
                         (!u.parent anchor) -- +(5pt,0pt) -- (.child anchor)
                         \forestoption{edge label};
                    }
                    ]
                ]
                [Search, t1node
                    [{E.g., Beam Search~\citeyearpar{geodrl}, MCTS~\citeyearpar{fgeo-drl}, PRS~\citeyearpar{visuothink}}, x2node,
                    ]
                ]
                [X-of-Thought, t1node
                    [{E.g., CoT~\citeyearpar{reprompting}, PoT~\citeyearpar{FUNCODER}, MCoT~\citeyearpar{MS-COT}}, x2node
                    ]
                ]
            ]
        ]
        [Training(§\ref{train}), onode 
            [Reinforcement Learning, tnode
                [LLM Alg., t1node
                    [{E.g., DPO~\citeyearpar{redstar}, GRPO~\citeyearpar{reason-rft}, GPG~\citeyearpar{gpg}}, x2node,
                    ]
                ]
                [Non-LLM Alg., t1node
                    [{E.g., DQN~\citep{geodrl}, PPO~\citeyearpar{dragon}}, x2node,
                    ]
                ]
            ]
            [Supervised Fine-Tuning, tnode
                [Data Filtering, t1node
                    [{E.g., ThinkLite-VL~\citeyearpar{ThinkLite-VL}, GeoGPT4V~\citeyearpar{geogpt4v}}, x2node
                    ]
                ]
                [Data Augmentation, t1node
                    [{E.g., GeoQA+~\citeyearpar{geoqa+_dpengs}, Geo170K~\citeyearpar{gllava}}, x2node
                    ]
                ]
                [Data Generation, t1node
                    [{E.g., GeomVerse~\citeyearpar{geomverse}, GeoGen~\citeyearpar{geogen_geologic}}, x2node,
                    ]
                ]  
            ]
            [Pre-Training, tnode
                [Pre-Training Data, t1node
                    [{E.g., AMPS~\citeyearpar{MATH_AMPS}, SynthGeo228k~\citeyearpar{DFE-GPS}}, x2node
                    ]
                ]
                [Pre-Training Task, t1node
                    [{E.g., MEP~\citeyearpar{unigeo}, JLP~\citeyearpar{geoqa_ngs}, PMP~\citeyearpar{lans}}, x2node,
                    ]
                ]
            ]
        ]
        [Archs. (§\ref{arch}), onode
            [Other Architectures, tnode
                [{E.g., GAN~\citeyearpar{GAN+CfER}, GNN~\citeyearpar{geodrl}, Decoder-Only~\citeyearpar{alphageometry}, Hybrid~\citeyearpar{pi-gps}}, x1node]
            ]
            [Encoder-Decoder Arch., tnode
                [Knowledge Module, t1node
                    [{E.g., Knowledge Extractor and Integrator~\citeyearpar{BiLSTM-CRF}, Theorem Predictor~\citeyearpar{gcn-fl}, Answer Verifier~\citeyearpar{PGPSNet-v2}}, x2node]
                ]
                [Decoder, t1node
                    [{E.g., LSTM~\citeyearpar{DualGeoSolver}, GRU~\citeyearpar{lans}, LLM~\citeyearpar{geodano_geoclip}}, x2node]
                ]
                [Fusion Module, t1node
                    [{E.g., co-attention~\citeyearpar{geoqa_ngs}, MLP~\citeyearpar{DFE-GPS}}, x2node]
                ]
                [Diagram Encoder, t1node
                    [{E.g., CNN~\citeyearpar{PGPSNet123}, ViT~\citeyearpar{SCA-GPS}, VQ-VAE~\citeyearpar{unimath}}, x2node]
                ]
                [Text Encoder, t1node
                    [{E.g., LSTM~\citeyearpar{geoqa_ngs}, GRU~\citeyearpar{geometryqa_s2g}, Transformer~\citeyearpar{DenseNet_gnsdtif}}, x2node]
                ]
            ]
        ]
    ]
    \end{forest}
    \caption{Taxonomy of deep learning methods for geometry problem solving.} 
    \vspace{0cm}
    \label{method_figure}
\end{figure*}

\subsection{Architectures for Geometry Problem Solving}
\label{arch}

In GPS, the classic deep learning architecture is the Encoder-Decoder architecture~\citep{encoder_decoder_seq2seq}, which also encompasses the widely used MLLMs in recent years. Other architectures have also been explored, including Generative Adversarial Networks (GANs)~\citep{gan}, Graph Neural Networks (GNNs)~\citep{gnn}, and Decoder-Only architectures. These architectures are outlined in more detail in Table~\ref{network_table}.

\subsubsection{Encoder-Decoder Architecture}

The encoder-decoder architecture can be divided into the following five key parts: text encoder, diagram encoder, multimodal fusion module, decoder, and optional knowledge module.

\noindent\textbf{Text Encoder.}
Text encoder can convert the text content of the geometry problem into formalized statements or encode it into vectors, enabling deep learning systems to process the text information.
Early studies usually use Long Short-Term Memory network (LSTM)~\citep{LSTM}, Gated Recurrent Unit (GRU)~\citep{GRU} and their bidirectional variants as text encoders, while more recent work employs Transformers~\citep{transformer} or pre-trained language models.

\noindent\textbf{Diagram Encoder.}
Parsing geometric diagrams into formal statements or encoding them into vector information is of great significance for solving multimodal geometry problems.
Early studies mainly used various Convolutional Neural Networks (CNNs)~\citep{CNN} to encode geometric diagrams, while recent studies have widely used pre-trained diagram encoders~\citep{vit, clip}. In addition, there are also studies that use LSTM, GNN, and other structures for diagram parsing.

\noindent\textbf{Multimodal Fusion Module.}
For multimodal geometry problems, the multimodal fusion module fuses and aligns text and diagram information extracted from the original problem or encoders, then passes it to the decoder.
Some works use a co-attention module~\citep{coattention} for multimodal fusion, and in MLLMs, structures such as MLP~\citep{llava1.5} are widely used. Additionally, some studies treat this module together with the decoder as a unified encoder-decoder architecture.

\noindent\textbf{Decoder.}
This module decodes the geometric knowledge and information to output the final answer to the question.
Many studies use LSTM or GRU as the decoder of deep learning systems. In addition, there are also a lot of studies using pre-trained LLMs.

\noindent\textbf{Knowledge Module.}
Some GPS systems integrate knowledge modules based on deep neural networks to more efficiently retrieve and apply knowledge and theorems in the field of geometry and verify the correctness of solutions.
The knowledge modules can be mainly divided into three categories: the first is \textit{Knowledge Extractor and Integrator}, which is used to extract and integrate geometric knowledge~\citep{DualGeoSolver}; the second is \textit{Theorem Predictor}, which is used to predict the geometric theorems required for the current solution step~\citep{gcn-fl}; and the third is \textit{Answer Verifier}, which is used to ensure the correctness of the solution~\citep{geogen_geologic}.

More details about the encoder-decoder architecture can be found in Appendix~\ref{en-de}.

\subsubsection{Other Architectures}

In addition to the encoder-decoder architecture, some deep learning systems for solving geometry problems have adopted other architectures.
\citet{GAN+CfER} adopt \textbf{GAN} architecture, while \citet{geodrl} and \citet{HGR} use \textbf{GNN} to solve geometry problems.
Many studies adopt \textbf{Decoder-Only Architecture}, for example, the AlphaGeometry series~\citep{alphageometry, Wu+AlphaGeometry, AlphaGeometry2} uses a trained Transformer to solve IMO geometry problems, and some work directly uses LLMs to solve unimodal geometry problems~\citep{dartmath,mathscale}.
Other studies have combined LLMs with other deep learning architectures~\citep{pi-gps, geouni}, or multiple LLMs~\citep{strategyllm, macm, dmad}, to build \textbf{Hybrid Architectures} for GPS.

\subsection{Training Stage for Geometry Problem Solving}
\label{train}

\subsubsection{Pre-Training}
\noindent\textbf{Pre-Training Task.} 
Beyond directly applying pre-trained models to geometry problems, many works design targeted pre-training tasks to enhance performance.
Some focus on the \textit{textual modality}: \citet{unigeo} propose Mathematical Expression Pretraining (MEP) to capture mathematical knowledge, while \citet{PGPSNet123}, \citet{PGPSNet-v2} and \citet{DenseNet_gnsdtif} adopt Masked Language Modeling (MLM) to improve the understanding and generation of textual descriptions.
Others target \textit{diagram encoders}: \citet{geoqa_ngs} introduce Jigsaw Location Prediction (JLP) and Geometry Elements Prediction (GEP), while \citet{SCA-GPS} apply Masked Image Modeling (MIM) and Multi-Label Classification (MLC) to optimize the diagram encoder.
There are also tasks focusing on \textit{matching multimodal relationships}, such as LANS~\citep{lans} with Structural-Semantic Pretraining (SSP) and Point-Match Pretraining (PMP), and SANS~\citep{sans} with Dual-Branch Visual-Textual Points Matching (DB-VTPM).

\noindent\textbf{Pre-Training Data.} To address the scarcity of geometric pre-training data, AMPS~\citep{MATH_AMPS} and InfiMM-WebMath-40B~\citep{InfiMM-WebMath-40B} offer large-scale mathematical and multimodal datasets, boosting model performance on geometry tasks. Given the gap between real-world images and geometric diagrams, some works construct dedicated datasets for diagram encoder pre-training. Geo-ViT~\citep{geox} compiles 120K+ diagrams for ViT training; CLIP-Math~\citep{mavis}, GeoCLIP~\citep{geodano_geoclip}, GeoGLIP~\citep{GeoGLIP}, and DFE-GPS~\citep{DFE-GPS} use synthetic data for geometry-focused visual pretraining.

\subsubsection{Supervised Fine-Tuning}

In GPS, deep learning models typically require Supervised Fine-Tuning (SFT), where training data plays a key role. In addition to collecting data from textbooks, exams, and the Internet, many studies focus on data generation, augmentation, and filtering of training data.

\noindent\textbf{Data Generation.} To obtain the geometric SFT data for training, various studies have employed different methods to accomplish \textit{geometric data synthesis}. \textit{Rule-based} approaches synthesize geometry problems using predefined generators~\citep{Data_Generator_for_Korean_Problem, alphageometry, visonlyqa, cogalign} or program templates that build complex diagrams from basic elements~\citep{geomverse, mavis, mathglance}. Recent studies further generate high-quality question-answer pairs with reasoning steps by multi-component pipelines~\citep{geogen_geologic, trustgeogen}. \textit{LLM-based} methods generate questions based on math concepts~\citep{mathscale, KPMATH}, with frameworks like GeoUni~\citep{geouni} and hybrid strategies combining rule-based image generation with LLM-based QA synthesis~\citep{R-COT}. \textit{Agent-based} approaches are also emerging~\citep{vista, wen2025feynman}, including competition-grade problems from Tonggeometry~\citep{tonggeometry}.

\noindent\textbf{Data Augmentation.} To improve robustness, many works apply rule-based augmentation to diversify text and diagrams~\citep{geoqa+_dpengs, PGPSNet123, gaps, dragon, sans, Math-puma}, use geometry theorems to create new problems~\citep{formalgeo, e-gps}, or adopt LLMs to generate diverse QA pairs~\citep{dartmath, Math-LLaVA, geovqa, GPSM4K2_mmasia, r3v}.
In addition, reasoning ability is enhanced by adding annotated \textit{reasoning traces}, including CoT~\citep{m3cot, gllava, sun2025mmverify, ursa, gns, vision-r1}, PoT~\citep{multilingpot, geocoder}, and long CoT~\citep{llava-cot, redstar, virgo, atomthink}. Other works improve geometric understanding by generating aligned \textit{diagrams} for unimodal geometry problems~\citep{bba, geogpt4v} or incorporating \textit{diagram descriptions} such as literals and captions~\citep{understanding_how_vlm, DFE-GPS, geox, autogeo}.

\noindent\textbf{Data Filtering.}
\citet{sun2025mmverify}, \citet{trustgeogen} and \citet{ThinkLite-VL} use search algorithms to screen data quality and difficulty, while \citet{geogpt4v}, \citet{InfiMM-WebMath-40B}, \citet{ursa}, \citet{visualwebinstruct} and \citet{KPMATH} use LLMs to score samples to screen for high-quality data.

\subsubsection{Reinforcement Learning}
Reinforcement Learning (RL) can significantly improve the geometric reasoning capabilities of deep learning models.

\noindent\textbf{Non-LLM Algorithms.}
Some studies have used Deep Reinforcement Learning (DRL) methods without LLMs to solve geometry problems~\citep{fgeo-drl}, such as the Deep Q-Network (DQN)~\citep{DQN} algorithm~\citep{geodrl,HGR} and the Proximal Policy Optimization (PPO)~\citep{ppo} algorithm~\citep{dragon}.

\noindent\textbf{LLM Algorithms.}
In LLM-based approaches, RL is typically introduced after SFT. Common algorithms include PPO~\citep{peng2024multimath, lmm-r1}, Direct Preference Optimization (DPO)~\citep{dpo, mavis, redstar, cogalign}, Group Relative Policy Optimization (GRPO)~\citep{deepseekr1, openvlthinker, reason-rft, curr-reft, VLAA-Thinking, noisyrollout, oneshot_RLVR, Hint-GRPO, vision-r1}, and Group Policy Gradient (GPG)~\citep{gpg}.

\subsection{Inference Stage for Geometry Problem Solving}
\label{infer}

\subsubsection{Test-Time Scaling}
Test-Time Scaling (TTS) has recently gained attention for significantly enhancing model reasoning during inference.

\noindent\textbf{X-of-Thought.}
X-of-Thought methods encourage LLMs to produce longer, more diverse outputs, which consume more computational resources than generating only short samples~\citep{tts_survey}.
Many works adopt different CoT~\citep{cot_origin} for GPS~\citep{reprompting, hsp, nullshot}, some of which involve multiple rounds of interaction with the model~\citep{php, CR}.
To boost arithmetic accuracy, PoT~\citep{pot_origin} is used to generate complete programs~\citep{mathsensei, FUNCODER} or distributed subprograms~\citep{sbsc}. Some studies combine CoT and PoT~\citep{xot, R&E}, or integrate CoT with external tools~\citep{creator,tora}. In addition, multimodal CoT approaches generate formal~\citep{Formal_Language_Generation} or natural language~\citep{understanding_how_vlm, VCAR, beyond_captioning} diagram descriptions before reasoning.

\noindent\textbf{Search Methods.} Many deep learning systems for GPS integrate tree-based search algorithms to enhance robustness, including Beam Search~\citep{geodrl, alphageometry, gaps, AlphaGeometry2, llava-cot}, Monte Carlo Tree Search (MCTS)~\citep{mcts_origin, fgeo-drl, mc_nest, mulberry, AR-MCTS, astar}, and Predictive Rollout Search (PRS)~\citep{visuothink}. Graph search is also explored~\citep{swap}.

\noindent\textbf{Verification Methods.}
A reliable verification method is crucial in TTS. Process Reward Models (PRMs) assess reasoning quality and often guide search paths~\citep{ursa, visualprm, vilbench, AR-MCTS, prm-bas, atomthink}. Other methods include using logits-based confidence~\citep{leco} or training an outcome verifier~\citep{genrm}.

\noindent\textbf{Others.}
\citet{r3v} use an LLM to select correct answers from multiple generated candidate solutions, while \citet{jin2025TTSforMAS} propose an agent framework to manage multiple agents and their reasoning strategies dynamically.

\subsubsection{Knowledge-Augmented Inference}

Knowledge-augmented inference enhances reasoning by incorporating external knowledge sources.

\noindent\textbf{Few-shot Learning.}
Few-shot learning~\citep{few-shot} guides models in solving similar geometry problems.
Several studies provide examples through In-Context Learning (ICL)~\citep{Hinting,comt}, some of which provide examples based on basic skills~\citep{skic}, some incorporate curriculum learning methods~\citep{cds}, and some place text in images~\citep{I2L}.
Others follow the Retrieval-Augmented Generation (RAG) paradigm to retrieve similar examples as hints~\citep{Geo-LLaVA, GPSM4K2_mmasia, geocoder}.

\noindent\textbf{Visual Aids.}
For GPS, some studies process the corresponding geometric diagrams during the inference stage to help solve the problem.
\citet{Vision-Augmented_Prompting} use drawing tools to convert text problems into multimodal input for reasoning, while some studies draw auxiliary lines or highlight key features on diagrams~\citep{SKETCHPAD, interactive_Sketchpad, cogcom, visuothink}.

\noindent\textbf{Others.}
\citet{learning2plan} employ learned task plans to guide reasoning, and \citet{em2} leverage explicit memory updates to utilize contextual knowledge captured during training.

%% file: section/eval.tex
\section{Evaluations for Geometry Problem Solving}
\label{eval}
In this section, we summarize the evaluation methods for GPS, including automatic and manual approaches.

\subsection{Automatic Evaluation}
Automatic metrics include performance-based metrics (outcome-based metrics and process-based metrics) and efficiency-based metrics.

\subsubsection{Performance-Based Metrics}
\noindent\textbf{Outcome-Based Metrics.}
Outcome-based metrics focus on measuring the accuracy of final answers without considering reasoning details.
\textit{Top-k Accuracy} (\textit{Top-k Acc}) and \textit{Pass@n} (\textit{P@n}) are two main metrics for answer accuracy, measuring the proportion of cases where a correct answer appears in the top \textit{k} predictions and the proportion of problems solved correctly at least once within \textit{n} attempts, respectively.
Other works also employ outcome-based metrics such as \textit{choice} (proportion of selecting the correct answer from multiple-choice options, or randomly if undetermined)~\citep{PGPSNet123}, \textit{F1 score} (considering both precision and recall)~\citep{numglue, comt}, \textit{maj@k} (proportion of obtaining the correct answer via majority vote among \textit{k} samples)~\citep{harp}, \textit{number of correct and wrong answers}~\citep{Number_Line_Problems}, and \textit{competition scores} such as SAT~\citep{geos} or IMO scores~\citep{alphageometry}.
Most metrics are evaluated using rule-based methods, with some adopting the ``LLM-as-a-Judge'' paradigm~\citep{llm-as-a-judge}.

\noindent\textbf{Process-Based Metrics.}
Recently, increasing attention has been paid to the reasoning process of deep learning systems, beyond just the final results, to further improve model performance.
To assess the executability of the reasoning process, \textit{Completion}~\citep{PGPSNet123} measures the accuracy of selecting the first executable solution, while \textit{No Result}~\citep{geoqa_ngs} indicates the ratio of cases where the reasoning program fails to produce output.
To evaluate the correctness of reasoning on benchmarks with standard CoT answers~\citep{GPSM4K2_mmasia, we-math}, some studies use metrics such as \textit{N-gram Similarity}~\citep{visaidmath}, \textit{Step Accuracy Rate}~\citep{MV-MATH}, and \textit{CoT-E score}~\citep{mathflow}, and extract step answers via rule-based methods or LLMs.
For other process-based metrics that are hard to quantify, such as step accuracy without reference CoT~\citep{mathverse, zhou2024mathscape, cmm-math} or logical coherence of CoT~\citep{DFE-GPS}, scoring is typically done with the help of LLMs.

\subsubsection{Efficiency-Based Metrics}
Efficiency-based metrics measure the model’s resource consumption and efficiency performance during reasoning, including the time required to solve the problem~\citep{GeoShader}, the failure rate within a time limit (\textit{timeout})~\citep{formalgeo}, the number of inference steps~\citep{e-gps, visuothink}, and the cost of running the model~\citep{mathconstruct}.

\subsection{Manual Evaluation}
Manual evaluation, which is rarely used in GPS, involves experts or annotators directly checking the model’s output or reasoning process.
Core uses include: (1) evaluating the correctness of complex answers (e.g., judging whether $\frac{1}{\sqrt{2}}$ equals $\frac{\sqrt{2}}{2}$)~\citep{conic10k}; (2) assessing the interpretability of the reasoning process~\citep{geos++, alphageometry}. Additionally, many studies manually check the reasons for wrong and correct answers, which is also called a case study~\citep{inter-gps, OlympiadBench}.

%% file: section/discuss.tex
\section{Discussion}
\label{discuss}

In this section, we provide an in-depth discussion of deep learning for GPS. We first analyze the performance of key models on mainstream benchmarks. Based on these findings and the broader literature, we then address key challenges and propose promising future directions.

\subsection{Performance Analysis}

\begin{table}[t]
\centering
\small
\setlength{\tabcolsep}{3pt}
\begin{tabular}{lccc}
\toprule
\textbf{Model} & \textbf{Geo3K} & \textbf{GeoQA} & \textbf{MathVista} \\
\midrule
\textbf{Inter-GPS}~\citeyearpar{inter-gps} & 57.5 & - & - \\
\textbf{NGS}~\citeyearpar{geoqa_ngs} & - & 60.7 & - \\
\textbf{LANS}~\citeyearpar{lans} & \textbf{82.3} & - & - \\
\textbf{SANS}~\citeyearpar{sans} & \underline{81.5} & - & - \\
\textbf{URSA-8B}~\citeyearpar{ursa} & - & 75.6 & \textbf{83.2} \\
\textbf{GeoUni-1.5B}~\citeyearpar{geouni} & 71.8 & \underline{78.0} & - \\
\textbf{SVE-Math-7B}~\citeyearpar{GeoGLIP} & - & \textbf{79.6} & 67.3 \\
\textbf{RedStar-Geo-8B}~\citeyearpar{redstar} & 33.6 & 64.0 & \underline{75.8} \\
\bottomrule
\end{tabular}
\caption{Performance summary of representative models on Geometry3K-test, GeoQA-test, and MathVista-testmini-GPS benchmarks (data from original papers).}
\vspace{-0.5cm}
\label{tab:performance_comparison}
\end{table}

Table~\ref{tab:performance_comparison} highlights the state-of-the-art and the second-best results on Geometry3K-test~\citep{inter-gps}, GeoQA-test~\citep{geoqa_ngs} and MathVista-testmini-GPS~\citep{mathvista} among over 40 models reviewed in this survey. 
Analysis reveals four trends: (1) \textit{reinforcement learning} effectively enhances geometric reasoning, as evidenced by impressive results on GeoQA and MathVista~\citep{GeoGLIP, ursa}; (2) \textit{neural-symbolic methods} like LANS~\citep{lans} and SANS~\citep{sans} demonstrate superior performance on symbolic-oriented tasks like Geometry3K; (3) the success of URSA~\citep{ursa} underscores the critical role of \textit{high-quality, large-scale training data}; and (4) \textit{test-time scaling} methods show promising potential that warrants further exploration~\citep{redstar}.

However, some of these improvements stem from the evolution of multimodal models' foundational capabilities. As these advance, existing benchmarks may become saturated, requiring \textit{more challenging evaluation}. Furthermore, while larger models generally exhibit superior performance, their substantial computational requirements remain a deployment hurdle. In contrast, \textit{smaller models}, though currently less performant, still hold significant potential for specialized future development~\citep{geouni,lmm-r1}.

\subsection{Challenges}

\noindent\textbf{Data.} First, \textit{current GPS data have significant limitations}. In terms of task type, geometry theorem proving is seriously underrepresented compared to numerical calculation. In terms of geometry type, solid and analytic geometry are lacking relative to plane geometry. In terms of language type, the data is mostly in English and Chinese, with little in other languages.
Second, \textit{a large gap remains between synthetic data and real exam questions}. Although recent methods can generate large-scale synthetic data for training, their performance improvement is still limited~\citep{geogen_geologic,trustgeogen}, which highlights the need for methods to synthesize more realistic and effective data.
Additionally, \textit{most datasets lack annotations for intermediate steps and reasoning processes}~\citep{Math-LLaVA}, which future work should address. More discussion is in Appendix~\ref{data}.

\noindent\textbf{Evaluation.}
First, \textit{question types are monotonous}. Existing benchmarks mainly use multiple-choice questions for evaluation (see Table~\ref{basic_core_data_table}), allowing models to guess and compromising evaluation accuracy. Some works mitigate this by permuting options~\citep{mathbench} or by not providing candidate options~\citep{trustgeogen}, but these methods have not yet become widespread.
Second, \textit{there is no standard method for evaluating the reasoning process}. As the demand for model improvement grows, reasoning evaluation for GPS has gained attention. However, existing methods lack unified standards, and more precise criteria are needed to better identify and address model deficiencies~\citep{Dataset_Evaluation}.
Third, \textit{current benchmarks may lack robustness}, as model performance often varies under slight perturbations~\citep{GEOMREL_GEOCOT, MATHCHECK}. Additionally, \textit{some datasets may appear in training data}, compromising fair evaluation~\citep{Dataset_Evaluation}, underscoring the need for more authoritative evaluation methods.

Furthermore, current evaluations largely overlook the dimension of computational efficiency. In practice, there is often a trade-off between performance-based and efficiency-based metrics; improvements in the former frequently come at the expense of the latter~\citep{visuothink, mathconstruct}. However, existing GPS research has paid limited attention to efficiency, remaining disproportionately focused on driving up performance scores. Even among the few studies that have considered efficiency, only a small fraction have successfully improved performance without sacrificing it~\citep{e-gps}.

\noindent\textbf{Capability.}
Current deep learning systems still show notable deficiencies in solving geometry problems.
Given the multimodal nature of most problems, the model’s geometric \textit{visual perception} ability is crucial.
However, studies show that adding diagrams often lowers accuracy compared to using text alone~\citep{DFE-GPS, highschool_education}.
In multimodal settings, spatial perception of diagrams remains a major bottleneck limiting overall performance~\citep{MM-MATH, gepbench, visonlyqa, euclid_Geoperception}. Studies show that deep learning models struggle to detect~\citep{Handwritten_Data_Identification, geodano_geoclip} and perceive~\citep{GEOMREL_GEOCOT, visnumbench} geometric angles, and often fail to accurately recognize line lengths~\citep{sp-1, cogalign}. 
These weaknesses may stem from the one-dimensional nature of model architectures~\citep{AIME24_eval}, the limited resolution of visual encoders~\citep{ATB-NGS, fgeo-Parser_FormalGeo7K-v2}, and their training on natural images~\citep{geoclidean, geocoder}, all of which hinder performance on geometric figures.
Additionally, many models continue to struggle with \textit{arithmetic accuracy}. Some adopt symbolic or formal reasoning~\citep{gns, AlphaGeometry2}, while others use external computation modules to mitigate this limitation~\citep{R&E, PGPSNet-v2, maths_mns}. LLMs may also develop a ``\textit{mindset}'', such as defaulting to coordinate system construction~\citep{AIME24_eval}, which can fail when such strategies are inapplicable.

\subsection{Future Directions}

\noindent\textbf{Combination of Perception and Reasoning.}
Studies show that visual perception and reasoning errors are the primary causes of model failures~\citep{Dataset_Evaluation, MV-MATH}. While early efforts targeted reasoning improvements, recent research has shifted toward perception; however, effectively integrating both remains a key challenge.
These two aspects are not mutually exclusive but rather complementary. For example, \textit{better modality alignment tasks} can be designed for specialized visual encoders or modules to enhance reasoning; \textit{more efficient multimodal CoT methods} can be explored to achieve deeper integration of perception and reasoning.

\noindent\textbf{Specialized Reinforcement Learning.} RL offers advantages over SFT in terms of complex reasoning, annotation efficiency, and adaptability, but faces training stability issues and high computational costs. Current RL applications in GPS mainly rely on general mathematical reasoning frameworks, lacking reward mechanisms specifically tailored to geometric contexts. Future research could focus on \textit{designing GPS-specific RL strategies}. Potential approaches include: (1) \textit{PRMs} that score intermediate spatial inferences or proof steps to encourage logical reasoning; (2) \textit{tool-use rewards} that convert deterministic feedback from geometric constructors, algebraic verifiers, or symbolic provers into reward signals. Notably, training datasets designed for SFT may not be suitable for RL~\citep{VLAA-Thinking}, which calls for careful consideration of diversity and generalization.

\noindent\textbf{Use of Cognitive Pattern.}
Cognitive pattern is a comprehensive approach that simulates human cognitive processes in understanding and solving complex problems~\citep{cognitive_pattern, Integrated_System}. Originating from early problem-solving research, many GPS strategies mimicking human problem-solving have proven effective~\citep{Adding_Auxiliary_Lines, Automatically_Expanding_Predicates}, such as \textit{highlighting key information} in diagrams and texts; \textit{referencing diagram annotations}; \textit{adding auxiliary lines}, \textit{coordinate axes}, and other diagram elements to clarify geometric structures; \textit{applying relevant theorems and knowledge}; and \textit{using curriculum learning} to progressively enhance problem-solving ability. However, these methods remain underutilized in current deep learning systems and warrant further investigation.

\noindent\textbf{Educational System.}
Before the rise of deep learning, many systems and tools had already been developed for geometry education, such as automatic scoring~\citep{GTP_Assessment}, theorem discovery~\citep{GeoGebra_Discover}, and problem-solving systems~\citep{AnalyticalInk, Ontology-Controlled, linguistic, Inductive_and_Deductive_Reasoning, Point_Geometry_Identity}, aimed at supporting teaching and learning. However, in the deep learning era, intelligent systems for geometry education remain relatively scarce. Automated GPS is seen as a key direction for future intelligent education~\citep{suffi-GPSC}. While recent AI tools have shown progress in solving geometry problems, they still face challenges in becoming effective educational tools---such as limited multi-language support and insufficient visual interaction. Their real-world capabilities remain constrained, and dedicated educational agents are still rare, highlighting the urgent need for further research to tackle the complex demands of this field.

%% file: section/conclu.tex
\section{Conclusion}
\label{conclu}

In this paper, we present a comprehensive and systematic survey of GPS. We summarize the relevant tasks, deep learning methods, and evaluation approaches, benchmark the performance of representative models, and provide an in-depth analysis of the limitations of current data, evaluation, and model capabilities. Finally, we look forward to possible future research directions and highlight the broad scope for exploration in this field. This article aims to provide readers who are interested in this field with a comprehensive and practical resource to meet their research needs.

%% file: section/data.tex
\section{Geometry Problem Solving Datasets}
\label{data}

In this section, we further analyze various datasets for GPS. 
Table~\ref{basic_core_data_table} and Table~\ref{Composite_table} provide a comprehensive summary of these datasets related to GPS tasks from multiple perspectives, including dataset name, task type, geometry type, grade level, problem source, presence of images, language, question format, rationale availability, sizes of training, validation, and test sets, as well as open-source status; a check mark indicates open-source datasets with links to the corresponding resources.

\noindent\textbf{The current data for geometry theorem proving remains insufficient.} Existing academic research predominantly centers on geometric numerical calculations, whereas studies on geometry theorem proving are relatively limited, and relevant data resources are still lacking. 
Despite sharing many similarities in problem formulation and underlying mathematical concepts~\citep{unigeo}, proof problems and calculation problems have distinct characteristics and challenges. Therefore, both types of geometry problems deserve equal attention.

\noindent\textbf{The current data for solid geometry and analytic geometry remains insufficient.} Most datasets used in GPS tasks are concentrated in plane geometry, while data for other geometry types---such as solid geometry~\citep{Solid_Geometric_Calculation_Problems} and analytic geometry~\citep{conic10k}---remain limited. One study notes that existing solid geometry problems are often overly simple and regular~\citep{geosense}, with diagrams containing only basic visual elements and rarely involving complex geometric combinations, thereby restricting progress in this area. Even within plane geometry, high-quality evaluation datasets are still scarce.

\noindent\textbf{The current data sources remain limited.} While existing datasets are generally authentic and reliable, they are often small in scale. Recently, due to the shortage of real-world data and concerns over copyright, many large-scale datasets have been constructed via data augmentation or programmatic synthesis~\citep{gllava, geogen_geologic}. However, the synthetic data often falls short in terms of realism, diversity, and quality, making it difficult to serve as a full substitute for real data.

\noindent\textbf{The current data coverage of language and question types remains limited.} In terms of language, existing datasets primarily cover English and Chinese, while authentic data involving other native languages~\citep{m3exam, m3gia} remains notably limited. This limits evaluation in the context of various national exams and reduces fairness. In terms of question types, most are multiple-choice, which allows models to guess answers and impairs accurate assessment of model reasoning ability.

\noindent\textbf{The current datasets remain lacking in rationale annotations.} Most datasets do not provide detailed annotations of intermediate reasoning steps~\citep{Math-LLaVA}. Even when rationales are included, they often lack standardized formatting and sufficient granularity, falling short of the needs for evaluating step-by-step reasoning. Moreover, the rationale annotations are typically presented in natural language, which may not meet the needs of deep learning systems that operate in formal languages.

\setlength{\tabcolsep}{2pt}
\begin{table*}[]
\centering
\fontsize{6.47pt}{8pt}\selectfont
\begin{tabular}{lccccccccccc}
\toprule
\textbf{Dataset}                                                           & \textbf{Task} & \textbf{Type}    & \textbf{Grade} & \textbf{Source}    & \textbf{Image}             & \textbf{Language} & \textbf{Question} & \textbf{Rationale} & \textbf{Trainval Size} & \textbf{Test Size}                                             & \textbf{Opensource}        \\
\midrule
\multicolumn{12}{c}{\cellcolor{gray!10}\textit{\textbf{Fundamental Tasks}}}\\
\midrule
\textbf{GeoC50}~\citeyearpar{GeoC50}                                                             & RE            & \textbf{P}       & -              & existing (dataset)              & \ding{51} & zh                & FR                & -                  & -                      & 50                                                             & \ding{55} \\
\textbf{2Dgeometricshapes}~\citeyearpar{2Dgeometricshapes}                                                  & ER            & \textbf{P}       & -              & program            & \ding{51} & -                 & CQ                & -                  & 36000 & 54000                                                               & \clickablecheckmark{https://data.mendeley.com/datasets/wzr2yv7r53/1} \\
\textbf{GeoRE}~\citeyearpar{geore}                                                              & RE            & \textbf{P}       & 6-12      & Internet           & \ding{55} & zh                & FR                & -                  & 10000                  & 2901                                                           & \clickablecheckmark{https://mathai4ed.github.io/papers/papers/paper_6_GeoRE_sample.json} \\
\bottomrule
\end{tabular}
\caption{A summarization of geometry problem solving datasets for fundamental tasks and core tasks. Task: ER: geometric element recognition, SR: geometric structure recognition, DP: geometric diagram parsing, DC: geometric diagram captioning, SP: semantic parsing for geometry problem texts, RE: geometric relation extraction, KP: geometric knowledge prediction, TP: geometry theorem proving, NC: geometric numerical calculation. Type: \textbf{P}: plane geometry, \textbf{S}: solid geometry, \textbf{A}: analytic geometry. Question: MC: multiple-choice, NR: numerical response, FR: free-response, FB: fill-in-the-blank, YN: yes-or-no, SA: short-answer, CQ: classification question. Rationale: nl: natural language. $^{\ast}$ indicates that the dataset contains more than just geometry-related content.}
\label{basic_core_data_table}
\end{table*}


\begin{table*}[]
\centering
\fontsize{6.6pt}{8.6pt}\selectfont
\begin{tabular}{lccccccccccc}
\toprule
\textbf{Dataset}                                                           & \textbf{Task} & \textbf{Type}    & \textbf{Grade} & \textbf{Source}    & \textbf{Image}             & \textbf{Language} & \textbf{Question} & \textbf{Rationale} & \textbf{Trainval Size} & \textbf{Test Size}                                             & \textbf{Opensource}        \\
 \midrule
\multicolumn{12}{c}{\cellcolor{gray!10}\textit{\textbf{Composite Tasks}}}                                                                                                                                                    \\
\midrule
\textbf{MATH}~\citeyearpar{MATH_AMPS}                & MR & \textbf{P}, \textbf{S}    & 9-12  & exam             & \ding{55} & en       & NR         & nl     & 7500        & 5000      & \clickablecheckmark{https://github.com/hendrycks/math} \\
\textbf{AMPS}~\citeyearpar{MATH_AMPS}                & MR & \textbf{P}, \textbf{S}    & 1-12  & website, program & \ding{55} & en       & NR, FR     & nl     & $\sim$5100k & -         & \clickablecheckmark{https://github.com/hendrycks/math} \\
\textbf{NumGLUE}~\citeyearpar{numglue}             & MR & \textbf{P}       & 6-10  & existing            & \ding{55} & en       & CQ         & -      & 81466       & 10583     & \clickablecheckmark{https://github.com/allenai/numglue} \\
\textbf{Lila}~\citeyearpar{mishra2022lila}       & MR & \textbf{P}, \textbf{S} & - & existing & \ding{55} & en & MC, FB, FR & program & 107052       & 26763 & \clickablecheckmark{https://huggingface.co/datasets/allenai/lila} \\
\textbf{DMath}~\citeyearpar{DMATH}                                                              & MR            & \textbf{P}       & 1-6       & handcraft        & \ding{55} & en, kr            & NR                & program            & 7943                   & 2079                                                           & \clickablecheckmark{https://github.com/JiwooKimAR/dmath} \\
\bottomrule
\end{tabular}
\caption{A summarization of geometry problem solving datasets for composite tasks. Task: MR: mathematical reasoning. Type: \textbf{P}: plane geometry, \textbf{S}: solid geometry, \textbf{A}: analytic geometry, \textbf{D}: differential geometry. Question: MC: multiple-choice, NR: numerical response, FR: free-response, FB: fill-in-the-blank, YN: yes-or-no, SA: short-answer, CQ: classification question. Rationale: nl: natural language.}
\label{Composite_table}
\end{table*}

%% file: section/other.tex
\section{Other Geometry Tasks}
\label{other}
In addition to GPS, some other geometry-related tasks, which have similar fundamental tasks, have not been systematically summarized. More details of the corresponding datasets can be found in Table~\ref{other_task_table}.

\begin{table*}[t]
\centering
\fontsize{7.5pt}{9.6pt}\selectfont
\begin{tabular}{lccccccccccc}
\toprule
\textbf{Dataset}                                                           & \textbf{Task} & \textbf{Type}    & \textbf{Grade} & \textbf{Source}    & \textbf{Image}             & \textbf{Language} & \textbf{Question} & \textbf{Rationale} & \textbf{Trainval Size} & \textbf{Test Size}                                             & \textbf{Opensource}        \\
\midrule
\multicolumn{12}{c}{\cellcolor{gray!10}\textit{\textbf{Other Geometry Tasks}}}                                                         \\
\midrule
\textbf{GMBL}~\citeyearpar{GMBL}&	TD &	\textbf{P}&	-	&exam	&\ding{55}&	en	&GD&	-&	-&	39&	\clickablecheckmark{https://github.com/rkruegs123/geo-model-builder?tab=readme-ov-file} \\
\textbf{LeanEuclid}~\citeyearpar{LeanEuclid}                                                         & AF            & \textbf{P}       & -              & existing, textbook    & \ding{51} & en                & FR                & -                  & 140                    & 33                                                             & \clickablecheckmark{https://github.com/loganrjmurphy/LeanEuclid?tab=readme-ov-file} \\
\textbf{Euclidea}~\citeyearpar{Euclidea}                                                           & CP            & \textbf{P}       & -              & website            & \ding{55} & en                & FR                & nl   & -                      & 98                                                             & \ding{55} \\
\textbf{PyEuclidea}~\citeyearpar{Euclidea}                                                          & CP            & \textbf{P}       & -              & website            & \ding{55} & program           & FR                & -                  & -                      & 98                                                             & \clickablecheckmark{https://github.com/mirefek/py_euclidea} \\
\textbf{MagicGeoBench}~\citeyearpar{magicgeo}                                                      & TD            & \textbf{P}       & 6-12      & exam               & \ding{55} & en                & GD                & -                  & -                      & 220                                                            & \ding{55} \\
\textbf{GeoX-pretrain}~\citeyearpar{geox} & DG & \textbf{P} & - & web, textbook & \ding{51} & - & GD & - & 127912 & - & \clickablecheckmark{https://huggingface.co/datasets/U4R/GeoX-data} \\
\bottomrule
\end{tabular}
\caption{A summarization of datasets for other geometry tasks. Task: TD: geometric text-to-diagram; CP: geometric construction problem; DG: geometric diagram generation; AF: geometric autoformalization. Type: \textbf{P}: plane geometry. Question: FR: free-response, GD: geometric diagram. Rationale: nl: natural language.}
\vspace{-0.5cm}
\label{other_task_table}
\end{table*}

\subsection{Geometric Diagram Generation} 
This task is dedicated to generating high-quality geometric diagrams. It aims to facilitate a deeper understanding of geometry problems and related applications such as image editing, thereby providing strong support for the field of education.

\noindent\textbf{Geometric Diagram Reconstruction.} 
This task has been the focus of some earlier works in the field of geometry.
It aims to use existing simple sketches or preliminarily drawn images to reconstruct a clearer and more standardized complete image, thereby helping users to understand and visualize the image content more intuitively~\mbox{\citep{Figure_Reconstruction_SDK}}.
One of the key challenges is to reconstruct 3D geometry from 2D single line drawing images~\citep{xue2010object,xue2012example,yang2013complex}, even if the input image is incomplete or inaccurate~\citep{zheng2015solid,zheng2016recovering,zheng2016context,zou2016example}.

\noindent\textbf{Geometric Text-to-Diagram.}
This task requires the system to be able to generate corresponding geometric diagrams from the natural language description of the geometry problem. 
This ability will significantly enhance the solution system's understanding, enabling it to more accurately interpret geometric propositions presented in flexible and diverse forms~\citep{G-ICS}. 
In addition to traditional rule-based methods~\citep{Image_Generation_for_Proofs, GMBL, alphageometry}, some recent studies have begun to use deep learning technology to build related systems~\citep{Text-to-Diagram, magicgeo, geouni}. 
MagicGeoBench~\citep{magicgeo} provides a dataset of 220 plane geometry problems from middle school mathematics exams, designed to evaluate the performance of text-to-diagram geometry generation models.

In addition to the above approaches, various other techniques have been developed for generating geometric diagrams.
Some tools, such as GeoGebra\footnote{\url{https://www.geogebra.org}} and Geometer’s Sketchpad~\citep{Geometers_Sketchpad}, support interactive constructions using virtual ruler and compass operations to generate geometric diagrams.
Additionally, non-interactive methods have also been proposed to automatically derive such constructions~\citep{bertot2004visualizing,itzhaky2013solving}.
To support more forms of geometric diagram generation, some studies have explored a wider range of methods to construct geometric diagrams. These methods include techniques like algebraic numerical optimization~\citep{gao2004mmp} and constrained numerical optimization~\citep{ye2020penrose}.

This task is also related to GPS. GeoX~\citep{geox} builds a pre-trained dataset containing more than 120,000 plane geometry images and tunes the visual encoder-decoder architecture using the mask auto-encoding scheme to obtain a visual encoder that fully understands geometric diagrams.
Additionally, some GPS work uses related methods to perform data enhancement on unimodal geometry problems and generate corresponding diagrams to obtain multimodal data~\citep{geogpt4v,bba, Vision-Augmented_Prompting}.

\subsection{Geometric Construction Problem}
Geometric construction problems, similar to problems in GPS, are also part of educational exams. 
Such tasks aim to use traditional ruler and compass construction methods to find an effective way to construct the desired figure.

In recent years, some studies have tried to use deep learning systems to solve geometric construction problems. 
In the online geometric construction game Euclidea\footnote{\url{https://www.euclidea.xyz}}, \citet{geometric_construction_problem} and \citet{euclidnet} use Mask R-CNN~\citep{maskrcnn} to solve difficult geometric construction problems using a purely image-based method. Additionally, \citet{Euclidea} convert the Euclidea problem into a Python format and solve it using a multi-agent framework based on LLMs. This provides us with new ideas and inspires us to further explore the application potential of deep learning systems in cognitive fields such as planning and auxiliary line addition.

\subsection{Geometric Figure Retrieval}
Before the widespread application of deep learning methods, the retrieval of plane geometry figures had always been an important topic in the field of scientific research~\citep{Bilayer-GAG, bos, Hierarchical_Searching, Plane_Geometric_Figures_Retrieval, Active_Learning, Learning2Rank}. With the advancement of computer technology, plane geometry retrieval may no longer be challenging in the era of deep learning. However, retrieving more complex solid geometry and irregular geometric figures may still be a direction worth studying.

\subsection{Geometric Autoformalization}
Autoformalization is a subtask of theorem proving~\citep{tp_survey}. A few studies focus on automatically converting informal geometry problems and proofs into formal theorems and proofs verifiable by machines. LeanEuclid~\citep{LeanEuclid} is a 173-problem geometric autoformalization dataset designed to test whether AI can understand mathematical problems and solutions written by humans and convert them into formal theorems and proofs.

%% file: section/encoder-decoder.tex
\section{Encoder-Decoder Architecture for Geometry Problem Solving}
\label{en-de}


In this section, we further elaborate on the deep learning components of the encoder-decoder architecture used for GPS. Table~\ref{network_table} provides a detailed summary of these components.

\subsection{Text Encoder}

Besides rule-based methods~\citep{inter-gps}, early research works typically use Recurrent Neural Networks (RNNs)~\citep{rnn} to parse~\citep{RSP123, Arsenal} or encode~\citep{geometryqa_s2g, geoqa_ngs} geometry problem texts. Common models include LSTM, GRU, and their bidirectional variants, BiLSTM and BiGRU. Some works employ Transformer~\citep{transformer} to encode text~\citep{DenseNet_gnsdtif, PGPSNet123, PGPSNet-v2}. Additionally, some research works use pre-trained language models for text encoding~\citep{Uniform_Vectorized_Syntax-Semantics, mcl, fgeo-Parser_FormalGeo7K-v2}, such as BERT~\citep{bert} and T5~\citep{T5}. Moreover, the dual encoder structure of RoBERTa~\citep{roberta} plus BiLSTM also shows good results~\citep{geoqa+_dpengs, SCA-GPS, DualGeoSolver, ATB-NGS}.

\subsection{Diagram Encoder}
Early studies primarily used CNNs to encode geometric diagrams~\citep{PGPSNet123, PGPSNet-v2, gold}, with network architectures including RetinaNet~\citep{retina_origin} and its DenseNet~\citep{densenet_and101_origin} variants~\citep{inter-gps, gcn-fl, RetinaNet+GCN, Uniform_Vectorized_Syntax-Semantics, DenseNet_gnsdtif}, ResNet~\citep{RESNET} and its ConvNeXt~\citep{convnet} variants~\citep{geoqa_ngs, geoqa+_dpengs, ATB-NGS, gaps}, and Fast R-CNN~\citep{fastrcnn, mcl}.
Recently, studies have widely adopted pre-trained diagram encoders, such as ViT~\citep{vit}, ViTMAE~\citep{VITMAE}, CLIP-ViT~\citep{clip}, SigLIP~\citep{SIGLIP}, and Swin-Transformer~\citep{swin_transformer}, primarily for building MLLMs.
Furthermore, \citet{Stacked_LSTM}, \citet{PGDPNet1}, and \citet{fgeo-Parser_FormalGeo7K-v2} use LSTM, GNN, and BLIP~\citep{li2022blip} to parse geometric diagrams, respectively, while UniMath~\citep{unimath} encodes diagrams through VQVAE~\citep{vqvae}.

Some other studies use a CNN-Transformer hybrid architecture to integrate the functions of a text encoder and a diagram encoder into a multimodal encoder~\citep{lans, sans}.

\subsection{Multimodal Fusion Module}
Drawing inspiration from \citet{coattention}, many studies introduce a co-attention module to comprehensively fuse and align text and image representations~\citep{geoqa_ngs, geoqa+_dpengs, SCA-GPS, DenseNet-121, DenseNet_gnsdtif}. Many MLLMs also incorporate multimodal fusion modules to enhance their multimodal understanding capabilities. 
For example, LLaVA-v1.5~\citep{llava1.5} and MAmmoTH-VL~\citep{mammothvl} both use a two-layer MLP visual-language connector~\citep{Math-LLaVA, eagle, Geo-LLaVA, geocoder, gns, visualwebinstruct}; GLM-4V~\citep{glm4v} and Qwen2.5-VL~\citep{qwen25technicalreport} use MLP to map image representations to text space~\citep{mathglm, geogen_geologic, lmm-r1}; and InternVL2~\citep{internvl2} uses the QLLaMA architecture~\citep{R-COT, redstar}. Additionally, some studies consider this module and the subsequent decoder as an overall encoder-decoder structure~\citep{mcl, PGPSNet123, unimath, lans, sans, gold}, employing self-attention units, BiGRU, and T5-Encoder.

\subsection{Decoder}
Many studies utilize LSTM~\citep{geoqa_ngs, geoqa+_dpengs, SCA-GPS, DenseNet-121, DualGeoSolver, ATB-NGS, DenseNet_gnsdtif} or GRU~\citep{geometryqa_s2g, mcl, PGPSNet123, lans, gaps, PGPSNet-v2} as decoders in deep learning systems, which may also integrate attention mechanisms.
Other studies employ pre-trained language models as decoders. For example, \citet{unimath} and \citet{gold} use the T5-Decoder, \citet{maths_mns} choose BERT, \citet{peng2024multimath} use DeepSeekMath-RL~\citep{deepseekmath}, and some studies use the Qwen series model~\citep{qwen} as the decoder~\citep{Geo-Qwen,Math-puma,euclid_Geoperception}. In addition, \citet{mavis} use MAmmoTH2~\citep{yue2024mammoth2}, \citet{DFE-GPS} choose Yi-1.5~\citep{young2024yi}, and \citet{geodano_geoclip} use Llama 3~\citep{llama3}.

\subsection{Knowledge Module}

\noindent\textbf{Knowledge Extractor and Integrator.} Some studies construct geometric knowledge frameworks using knowledge graphs. \citet{rmcee} and \citet{BiLSTM-CRF} use BiLSTM to extract geometric relationships, while \citet{geometryqa_s2g} embed knowledge graphs into vector space using Graph Convolutional Network (GCN)~\citep{gcn}. \citet{Geo-LLaVA} and \citet{geocoder} use CLIP and VISTA~\citep{zhou2024vista} models to encode geometry problems for retrieving similar problems. Additionally, \citet{DualGeoSolver} build a complete knowledge system through LSTM.

\noindent\textbf{Theorem Predictor.}
The theorem predictor is used to predict the geometry theorems needed for the current solution step to derive a formal solution path.
\citet{gcn-fl} and \citet{RetinaNet+GCN} encode the structural information of the formal language through GCN and use a BiLSTM-GRU based Sequence-to-Sequence (Seq2Seq) architecture~\citep{encoder_decoder_seq2seq} for theorem prediction.
In addition, many studies use a Transformer-based Seq2Seq architecture for prediction~\citep{inter-gps, e-gps, fgeo-HyperGNet}, and some introduce the T5 model~\citep{suffi-GPSC, fgeo-tp, Geo-Qwen}. 
Furthermore, \citet{fgeo-drl} leverage DistilBERT~\citep{distilbert} to guide the training of theorem predictors.

\noindent\textbf{Answer Verifier.}
Ensuring the correctness of the solution logic is one of the key steps in solving geometry problems. In addition to the traditional rule-based verification method~\citep{PGPSNet-v2}, \citet{geogen_geologic} introduce a pre-trained LLM~\citep{qwen25technicalreport} to verify the solution steps.